 \documentclass[conference, 10 pt]{IEEEtran}
\IEEEoverridecommandlockouts    

\usepackage{multirow}
\usepackage{lipsum}
\usepackage{amsmath}
\usepackage{amssymb}
\usepackage[ruled,vlined]{algorithm2e}
\usepackage{graphicx}
\usepackage{subcaption}
\usepackage{caption}
\usepackage{scalerel}
\usepackage{tikz}
\usetikzlibrary{svg.path}
\usepackage{pifont}
\usepackage{verbatim}
\usepackage{soul}
\usepackage[letterpaper,
            bindingoffset=0.2in,
            left=0.75in,
            right=0.75in,
            top=1in,
            bottom=0.75in,
            footskip=.25in]{geometry}
\usepackage{float}

\newcommand\x{0.065}
\newcommand\y{0.065}
\newcommand\z{0.065}
\newcommand{\Star}[1]{#1\ensuremath{^*}\kern-\scriptspace}

\definecolor{orcidlogocol}{HTML}{A6CE39}
\tikzset{
  orcidlogo/.pic={
    \fill[orcidlogocol] svg{M256,128c0,70.7-57.3,128-128,128C57.3,256,0,198.7,0,128C0,57.3,57.3,0,128,0C198.7,0,256,57.3,256,128z};
    \fill[white] svg{M86.3,186.2H70.9V79.1h15.4v48.4V186.2z}
                 svg{M108.9,79.1h41.6c39.6,0,57,28.3,57,53.6c0,27.5-21.5,53.6-56.8,53.6h-41.8V79.1z M124.3,172.4h24.5c34.9,0,42.9-26.5,42.9-39.7c0-21.5-13.7-39.7-43.7-39.7h-23.7V172.4z}
                 svg{M88.7,56.8c0,5.5-4.5,10.1-10.1,10.1c-5.6,0-10.1-4.6-10.1-10.1c0-5.6,4.5-10.1,10.1-10.1C84.2,46.7,88.7,51.3,88.7,56.8z};
  }
}

\newcommand\orcidicon[1]{\href{https://orcid.org/#1}{\mbox{\scalerel*{
\begin{tikzpicture}[yscale=-1,transform shape]
\pic{orcidlogo};
\end{tikzpicture}
}{|}}}}

\usepackage[bookmarks=false]{hyperref} 

\graphicspath{{./Figures/}}

\title{\LARGE \bf Assessing Localization Technologies for Pedestrian Collision Avoidance}

\author{
	\parbox{\textwidth}{%
		\centering
		Joshua Varughese$^{1}$\orcidicon{0000-0002-3250-0742} \emph{Member, IEEE}, Joseba Gorospe$^{1}$\orcidicon{0000-0003-0334-5509} \emph{Member, IEEE}, \\Novel Certad$^{1}$\orcidicon{0000-0002-5211-3598} \emph{Graduate Student Member, IEEE} and Cristina Olaverri-Monreal$^{1}$\orcidicon{0000-0002-5211-3598} \emph{Senior Member, IEEE}%
	}%
	\thanks{\textsuperscript{1}Dept.  Intelligent Transport Systems, Johannes Kepler University Linz, Altenberger Straße~69, 4040~Linz, Austria.
		{\tt\small \{joshua.varughese, joseba.gorospe, novel.certad\_hernandez, cristina.olaverri-monreal\}@jku.at}}%
}

\begin{document}

\maketitle              

\thispagestyle{empty}
\pagestyle{empty}

\begin{abstract}
 Robust pedestrian safety is crucial to the next-generation of intelligent transportation systems. Such systems rely on active pedestrian localization and predictive collision alerts. Pedestrian localization can be supported by Ultra-Wideband technology and Bluetooth 6.0, which offer high-precision ranging and low-latency communication, making them promising candidates for vehicular collision warning systems. This paper assesses the localization accuracy of these technologies for pedestrian alerting and benchmarks their performance against Global Navigation Satellite Systems. Experimental evaluations performed in this paper focused on key performance metrics, including localization accuracy and robustness to environmental conditions. Preliminary results suggest that Ultra-Wideband and Bluetooth 6.0 can serve as viable alternatives or complements to Global Navigation Satellite Systems in certain scenarios, improving situational awareness and enabling timely pedestrian alerts.


\end{abstract}
\section{Introduction} \label{sec:introduction}

\begin{figure}[htbp]
\centerline{\includegraphics[width=0.4\textwidth]{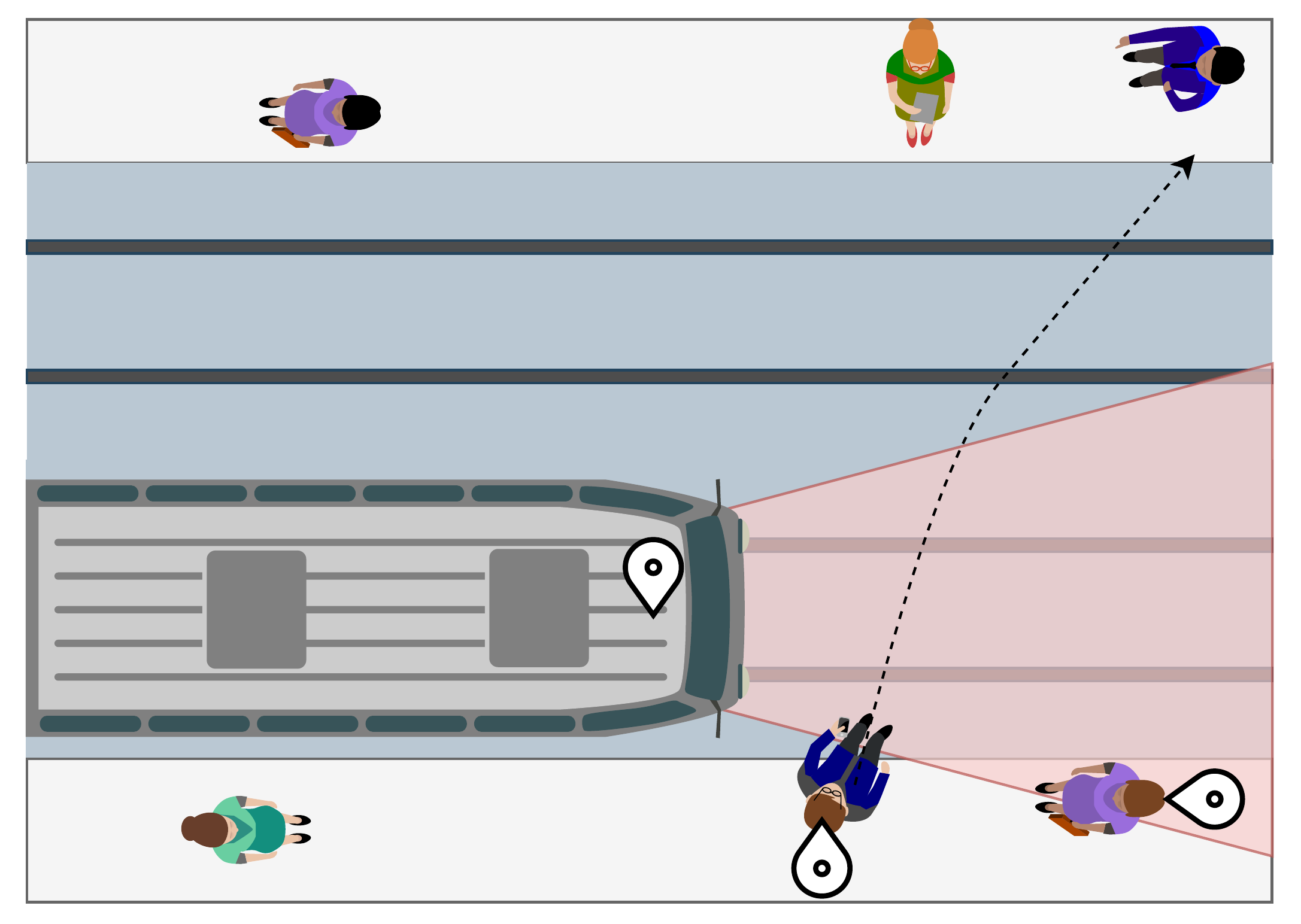}}
\caption{In this scenario, trams and VRUs share the road, and distracted VRU may cross diagonally without noticing an approaching tram. The vehicle must estimate the position of vulnerable road users within the region of interest (red).}
\label{fig:motivation}
\end{figure}

Road traffic accidents remain a major global safety concern, disproportionately affecting Vulnerable Road Users (VRUs), such as pedestrians and cyclists \cite{ELHAMDANI2020102856}. Pedestrians are among the most vulnerable road users. Unlike vehicle occupants, they are not protected by a vehicle structure or dedicated protective equipment (e.g., helmets) and are directly exposed to environmental conditions such as rain and sunlight \cite{monreal2016shadow}. Accidents involving VRUs often occur at intersections, bus stops, and areas with heavy urban traffic, where situational awareness is critical but often compromised. In addition, pedestrian distraction related to mobile phone use has emerged as a prominent safety concern, particularly in urban environments characterized by high traffic density \cite{Scharz2015pedestrian}. The risk is further amplified near buses and trams, as detailed in \cite{delre2025v2p}, particularly where infrastructure lacks safe crossings, sidewalks, or designated waiting areas. An exemplary scenario is illustrated in \autoref{fig:motivation}. Conventional safety measures, such as traffic lights, pedestrian crossings, and vehicle horns, are often insufficient to prevent accidents in these scenarios, as they rely on the VRU actively noticing and reacting to signals. 

Several approaches have been deployed to reduce collisions between vehicles and VRUs, ranging from infrastructural solutions to technology-driven interventions. Systems deployed in commercial vehicles include blind-spot warning systems, reversing sensors and cameras, and Automated Emergency Braking (AEB). In addition, smart pedestrian crossings equipped with sensor-activated LED lighting have been implemented to improve pedestrian visibility and driver awareness \cite{PORTERA2024crosswalk}. Although these systems have demonstrated effectiveness in reducing collisions, they primarily rely on timely perception and reaction both by pedestrians and the driver.

On the vehicular side, multiple sensors provide rich contextual information. However, for VRUs to react effectively to imminent hazards, this information must be communicated to them in a timely and reliable manner. Vehicle-to-Pedestrian (V2P) communication has therefore emerged as a promising approach to enhance situational awareness for VRUs. By enabling vehicles and pedestrians to exchange information such as position, speed, and trajectory, V2P systems can proactively warn both parties of potential collisions \cite{hussein2016p2v}. Several communication technologies can support this information exchange, including C-V2X or ITS-G5 \cite{abboud2016interworking}. Despite their potential, these systems still rely heavily on accurate localization systems, which are typically provided by the Global Navigation Satellite System (GNSS). In dense urban environments, GNSS-based positioning can suffer from reduced accuracy and availability \cite{zhu2018gnss}, potentially leading to delayed, missed, or incorrect warnings.

An emerging solution is Ultra-Wideband (UWB) technology, which offers a potential alternative by enabling both high-precision ranging and low-latency communication, particularly suitable in scenarios with weak or no GNSS signal. Although UWB has been successfully evaluated in static and indoor scenarios \cite{mazhar2017uwbindor, zhang2020pedindor}, its performance in outdoors, in vehicular environments where the position of both vehicles and VRUs is constantly changing, remains underexplored.

Another promising avenue to explore is the channel sounding technique included in Bluetooth 6.0. Although previous research has investigated the use of bluetooth based short range communication \cite{filgueiras2014sensing} to obtain new insights into mobility dynamics, Bluetooth Channel Sounding (BTCS) has not been explored in the context of P2V or V2P communication. Using channel sounding, localization accuracy can be improved by analyzing phase measurements across multiple RF frequencies. Unlike traditional localization based on Received Signal Strength Indicator (RSSI), channel sounding offers improved accuracy due to the phase based ranging method it employs. Given this technological advancement and the widespread implementation of Bluetooth technology on smartphones, this technology has the potential to significantly impact the field of Intelligent Transport Systems. Therefore, the BTCS feature of Bluetooth 6.0 is explored in this paper as a possible alternative for localizing and warning VRUs. 

Our contributions include the following aspects: 1) An experimental assessment of UWB and BTCS for V2P safety applications, 2) a comparison with Smart Phone-based Location (SPL) and a Real-Time Kinematic (RTK)-corrected GNSS signal used as a reference, 3) identification of their advantages and limitations in real-world scenarios, 4) discussion of their integration potential with public transport systems to enhance VRU safety. 

The remainder of this paper is organized as follows: \autoref{sec:related} reviews related work on localization mechanisms investigated in previous studies. Section \ref{sec:method} describes the communication modules, experimental setup and experiments performed in this work. Sections \ref{sec:results} and \ref{sec:discussion} present the results and  a discussion, respectively. Section \ref{sec:conclusion} concludes the paper.




\section{Related work}\label{sec:related}

Although prior research has largely emphasized pedestrian localization and tracking, only a small portion of the literature explicitly addressed pedestrian warning in hazardous situations. For example, \cite{rahimian2018harnessing} demonstrated the potential of explicit warnings to improve pedestrian safety. Similarly, studies \cite{hussein2016p2v, certad2025v2p, delre2025v2p} showed that pedestrian safety can be enhanced through explicit interaction with VRUs.


Accurate pedestrian tracking is a fundamental prerequisite for effective warning systems. Accordingly, research in this area focused on developing reliable methods to detect and follow pedestrian movements. Early work relied on handcrafted features \cite{dollar2014fast} and motion models, while more recent research increasingly used deep learning to improve robustness in crowded or occluded environments \cite{razzok2023pedestrian}. In \cite{picello2025leveraging}, the authors surveyed existing research and categorized localization and tracking methods based on the sensors used to locate the pedestrian. Various localization approaches have been proposed, including vision-based \cite{dollar2014fast, razzok2023pedestrian}, LiDAR-based \cite{gomez2023efficient,zhang2024robust}, RGB-D-based \cite{saha2022efficient}, thermal-based \cite{altay2022use}, and hybrid methods \cite{bellotto2008multisensor}; however, not all are suitable for deployment in road-user scenarios. For example, while thermal-based cameras might prove useful for remote areas with low footfall, data from such sensors are not helpful in a scenario where VRUs might have to be detected among a group of people. Furthermore, collecting vision and RGB-D images from public settings might require special authorizations due to regulations on image data due to data privacy. Additionally, while LiDAR-based localization of pedestrians offers a promising avenue, all image sensor based pedestrian tracking faces the same limitation of lacking a communication pathway. In such cases, communicating with the VRUs will remain a problem to be solved separately. Therefore, localization modalities which also enable communication, such as BTCS and UWB, need to be explored for V2P safety applications. 

The authors of \cite{park2024pedloc} implemented a pedestrian localization using UWB anchors mounted on a moving vehicle. Here, the authors developed a novel peak detection algorithm to improve the accuracy of detections. In \cite{kim2025deep}, the authors followed a similar approach using UWB anchors but used deep learning to locate pedestrians. Although these approaches are promising, there is no evaluation with other emerging modalities such as BTCS.

In \cite{hussein2016p2v}, the authors explored a communication and warning system based on the positions of pedestrians as measured by GNSS. While the application is promising, the authors identified the dependence on GNSS as a major limitation. Additionally, the authors employed the cellular network to communicate the warning to the VRU. Even though the cellular network is reliable, it is unsuited for safety critical applications such as collision avoidance due to possible communication delay.

Existing research indicated a clear need for a single-channel modality capable of explicitly locating and warning pedestrians. While promising approaches have been proposed \cite{park2024pedloc, hussein2016p2v, kim2025deep}, significant gaps remain that require further investigation and development. This paper evaluates emerging communication modalities, namely UWB and BTCS, and assesses their localization performance in a traffic scenario using a real robotic platform.





\begin{figure}[htbp]
    \centering
    \includegraphics[width=0.75\linewidth]{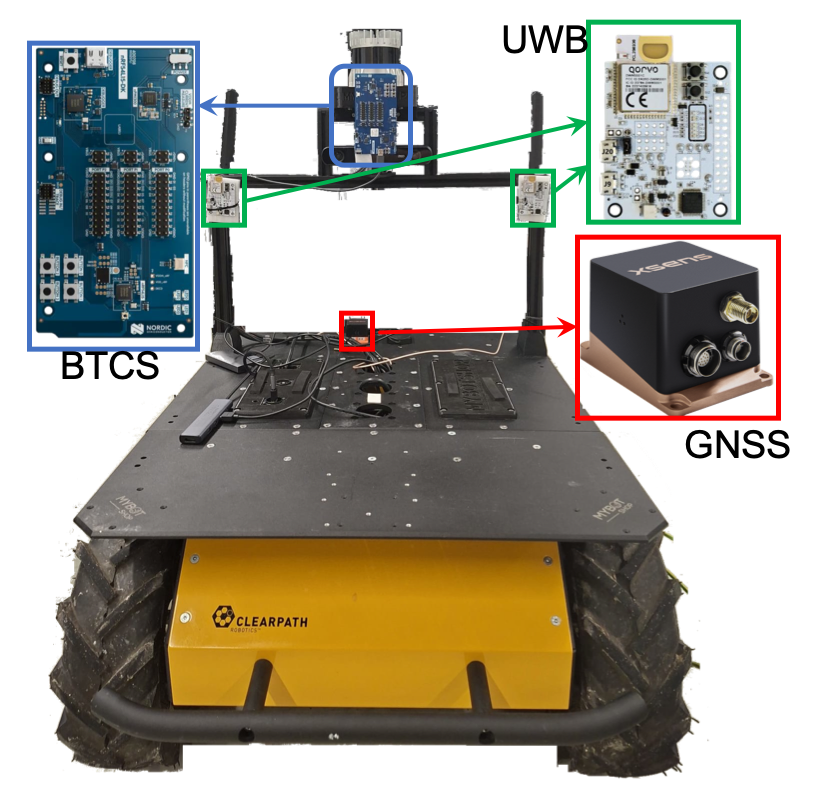}
    \caption{Integration of two UWB evaluation kit (green), a Bluetooth 6.0 evaluation kit, and the GNSS receiver on the clearpath A200 robot.}
    \label{fig:husky}
\end{figure}

\section{Method}\label{sec:method}

In this section, the methodology used for the evaluation of UWB, BTCS, and SPL against an RTK-corrected GNSS signal is described. A brief outline of the modules used is provided below. Furthermore, experiments were conducted in built-up and open areas, with or without Line Of Sight (LOS), and with or without antenna alignment to assess performance under varying conditions.

\subsection{Hardware modules} \label{sec:hardware}
This section describes the hardware modules integrated on the experimental platform for the performance evaluation. The system consists of Bluetooth 6.0, UWB, GNSS modules, as well as a commercial smartphone. The relevant components of the modules mentioned above were mounted on a Clearpath Husky A200 mobile robot, as shown in \autoref{fig:husky}.

\subsubsection{Bluetooth channel sounding (BTCS)} In order to conduct the evaluation of BTCS, two Bluetooth 6.0 enabled chips, nRF54L15-DK from Nordic Semiconductor were used. The distance between two nRF54L15-DK boards based on Time Difference of Arrival (TDoA) were measured.

\subsubsection{Ultra-Wideband (UWB)} As for evaluating UWB, three Qorvo DWM3001CDK boards were used. These boards are development kits designed to evaluate and prototype with the  fully integrated DWM3001C UWB module. 
By using three of these kits in appropriate modes, a communication initator could localize themselves with respect to the others using the Time Difference of Arrival (TDoA) principle.
 
\subsubsection{Global Navigation Satellite System (GNSS)}
An RTK-capable GNSS receiver was used to provide ground truth reference data with centimeter-level accuracy under ideal conditions. 

 \subsubsection{Smart Phone based Location (SPL)}
As the ultimate objective of this work was to compare emerging communication modalities with existing services, location data were also collected from a commercial smartphone. The SPL estimates are based on sensor fusion, combining data from GNSS, an inertial measurement unit, a gyroscope, and other integrated sensors.

\subsection{Experimental Setup}
\label{sec:setup}
All modules outlined in the previous section were integrated into a clearpath A200 ``Husky'' robot. The robot's middleware, Robot Operating System, (ROS2) was used as a hub to collect the data from the sensors. In order to synchronize the data, a Network Time Protocol (NTP) server was implemented on the robot, to which the other sensors could connect, thus minimizing time deviation across devices. In addition, SPL was collected from the smartphone using a web server running on the robot. On receiving a post request, the web server used a javascript library to inform ROS2 of these location messages.

\begin{figure}[htbp]
\centerline{\includegraphics[width=0.47\textwidth]{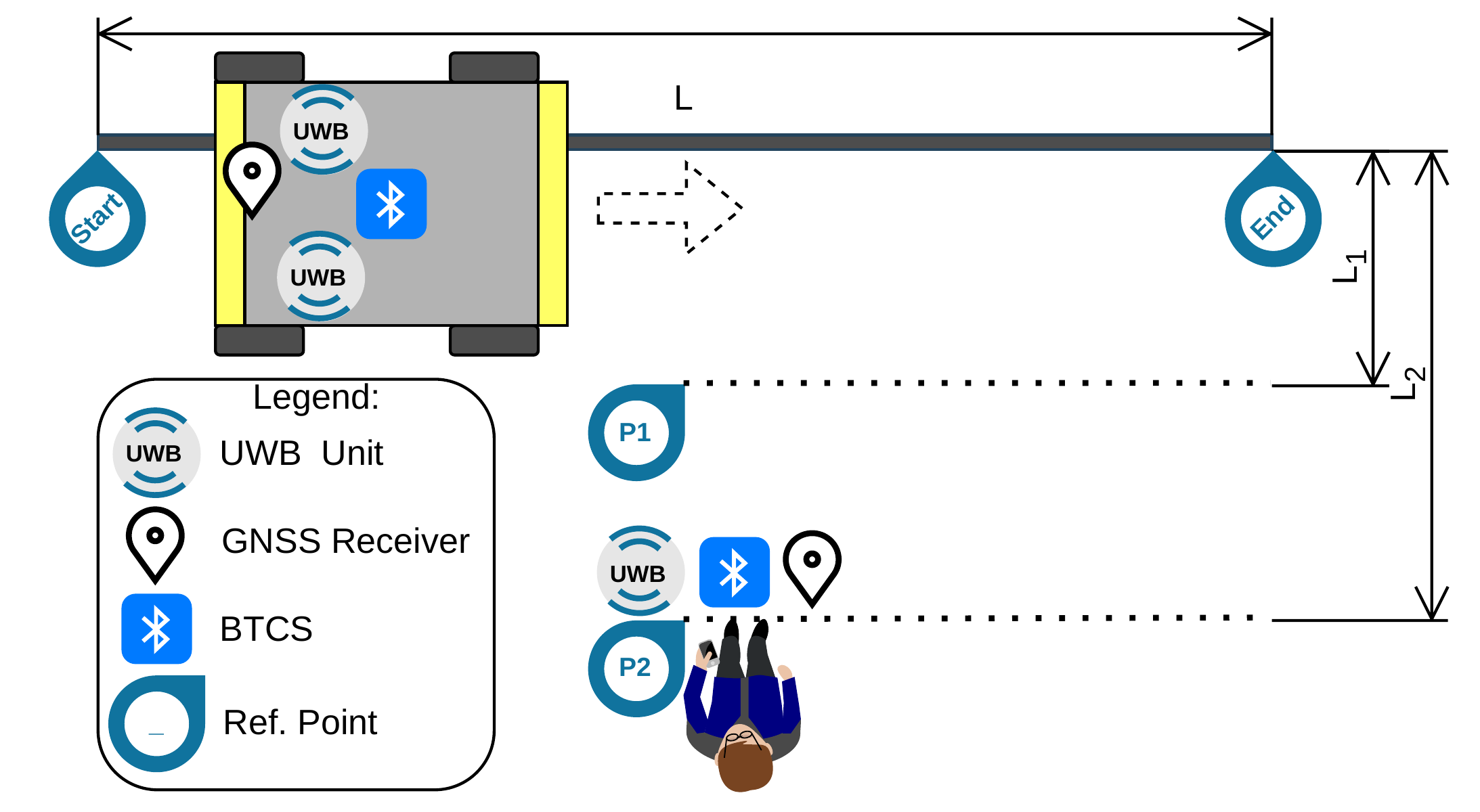}}
\caption{During the experiments, the position and distance of a pedestrian with respect to a robot was tracked using UWB, BTCS and a smart phone.}
\label{fig:experiment}
\end{figure}

As mentioned in \autoref{sec:introduction}, the main goal of this paper is to evaluate UWB, BTCS and GNSS for the purpose of alerting VRUs. Since pedestrian localization constitutes the first part of the problem, it was evaluated in a mock setup using the Husky robot. During the experiment, the robot moved along a straight-line trajectory ($L$) from ``Start'' to ``End'', as illustrated in \autoref{fig:experiment}. At the pedestrian position ($P_{1}$ and $P_{2}$), a UWB initiator (DWM3001CDK) and a BTCS initiator (nRF54L15-DK) were installed with their antennas oriented toward the ``Start'' position. In addition, a smart phone also was placed at the pedestrian position. Once the robot passed the pedestrian position, the UWB and BTCS initiator antennas were no longer aligned with the corresponding responder nodes mounted on the robot, resulting in a Non-Line-of-Sight (NLOS) orientation between the devices.



A total of 12 experiments were conducted in three distinct outdoor environments to assess system performance under varying environmental conditions: (i) A wide open area, (ii) an area close to a building, introducing potential multipath effects, and (iii) an area covered by a roof structure to partially attenuate satellite and radio signal propagation. These scenarios enabled evaluation under different levels of signal obstruction, reflection, and environmental complexity. In each environment, two pedestrian configurations ($P_{1}$ and $P_{2}$) were evaluated under both LOS and NLOS conditions. The LOS condition was obtained by aligning the communication devices to ensure unobstructed propagation, whereas the NLOS condition was introduced by positioning the pedestrian between the communication devices to obstruct the direct path.


\subsection{Evaluation}
Due to the presence of multiple coordinate systems, the evaluation was conducted using distances. Various implementation details of the experiment are outlined below: 

\subsubsection{Coordinate transformations}
Measurements based on Time Difference of Arrival (TDoA) such as UWB and BTCS resulted naturally in distance measurements rather than position measurements. GNSS data, however, was collected in the EPSG:4326 WGS 84 coordinate system. To enable comparison between these modalities, the GNSS coordinates were transformed to EPSG:3395 WGS 84 which enable simple distance calculations in 2D cartesian coordinate system. 
\subsubsection{Ground truth}
The ground truth positions throughout the experiment were obtained from a kalman filter fusing the encoder readings and the IMU onboard the robot. The starting point of the robot as measured by the RTK-corrected GNSS sensor was defined as a reference point of the local coordinate frame. Subsequent positions were computed by integrating the robot’s incremental motion in the x and y directions as provided by the wheel odometry.
\subsubsection{Performance measures}
To assess the performance of the different communication modalities, Root Mean Squared Error (RMSE) was adopted as defined below:

\begin{equation}
\text{RMSE} = \sqrt{\frac{1}{N} \sum_{i=1}^{N} (d_i - d_i^{\text{gt}})^2} 
\label{eq:rmse}
\end{equation}

where $N$ is the total number of measured distances, $d_i$ is the distance measured and $d_i^{gt}$ is the corresponding ground truth distance based on the odometry of the robot and the real position of the pedestrian.

In addition to the RMSE and standard deviation of the ranging error, the performance of the system was further evaluated by distance- and position-based analyses. 

First, the measured distances were plotted over time and compared with the corresponding ground truth distances. This temporal representation allowed identification of systematic deviations, transient errors, and environment-dependent effects. Second, pedestrian positions were estimated from the measured distances and visualized in the two-dimensional \(x\)-\(y\) coordinate frame. The \(+x\) axis was along the direction of movement of the robot. The rest of the axes followed the right hand rule according to REP-103 \cite{rep103}. This enabled the assessment of the overall localization performance, as the ultimate objective of the system was accurate position estimation rather than distance estimation alone. Please note that the estimation of the position of the pedestrian based on BTCS was not included due to a lack of anchors.


\section{Results}\label{sec:results}

As outlined in \autoref{sec:setup}, a set of 12 experiments were conducted to obtain the distances between the pedestrian and the robot using different modalities. The results are reported in \autoref{tab:open}, \autoref{tab:builtup} and \autoref{tab:roof} representing the three locations of experiments: open, built up and roofed areas respectively. For each experiment, the RMSE and standard deviation were computed, with all distance measurements expressed in meters.

\begin{table}[h]
\caption{Distance measurements in meters from various modalities in the experiment conducted in an open space}
\label{tab:open}
\resizebox{\linewidth}{!}{
\begin{tabular}{p{\z\linewidth}p{\z\linewidth}p{\z\linewidth}p{\z\linewidth}p{\z\linewidth}p{\z\linewidth}p{\z\linewidth}p{\z\linewidth}p{\z\linewidth}}
 & \multicolumn{2}{c}{$P_{1}$} & \multicolumn{2}{c}{$P_{2}$} & \multicolumn{2}{c}{$P_{1}$\textsuperscript{occ}} & \multicolumn{2}{c}{$P_{2}$\textsuperscript{occ}}\\
 & RMSE  & StD   & RMSE  & StD   & RMSE  & StD   & RMSE  & StD \\ 
\hline
SPL: & 14.678 & 4.724 & 2.879 & 2.264 & 2.534 & 2.335 & 1.63 & 1.557 \\ 
UWB: & 0.226 & 0.223 & 0.269 & 0.226 & 0.314 & 0.312 & 0.219 & 0.208 \\ 
BLE: & 0.718 & 0.247 & 1.022 & 0.722 & 1.236 & 1.194 & 1.49  & 1.489 \\ 
\hline
\end{tabular}
}
\end{table}

\begin{figure*}[h]
    \centering
         \begin{subfigure}[b]{0.44\linewidth}
        \centering
                \includegraphics[width=\textwidth]{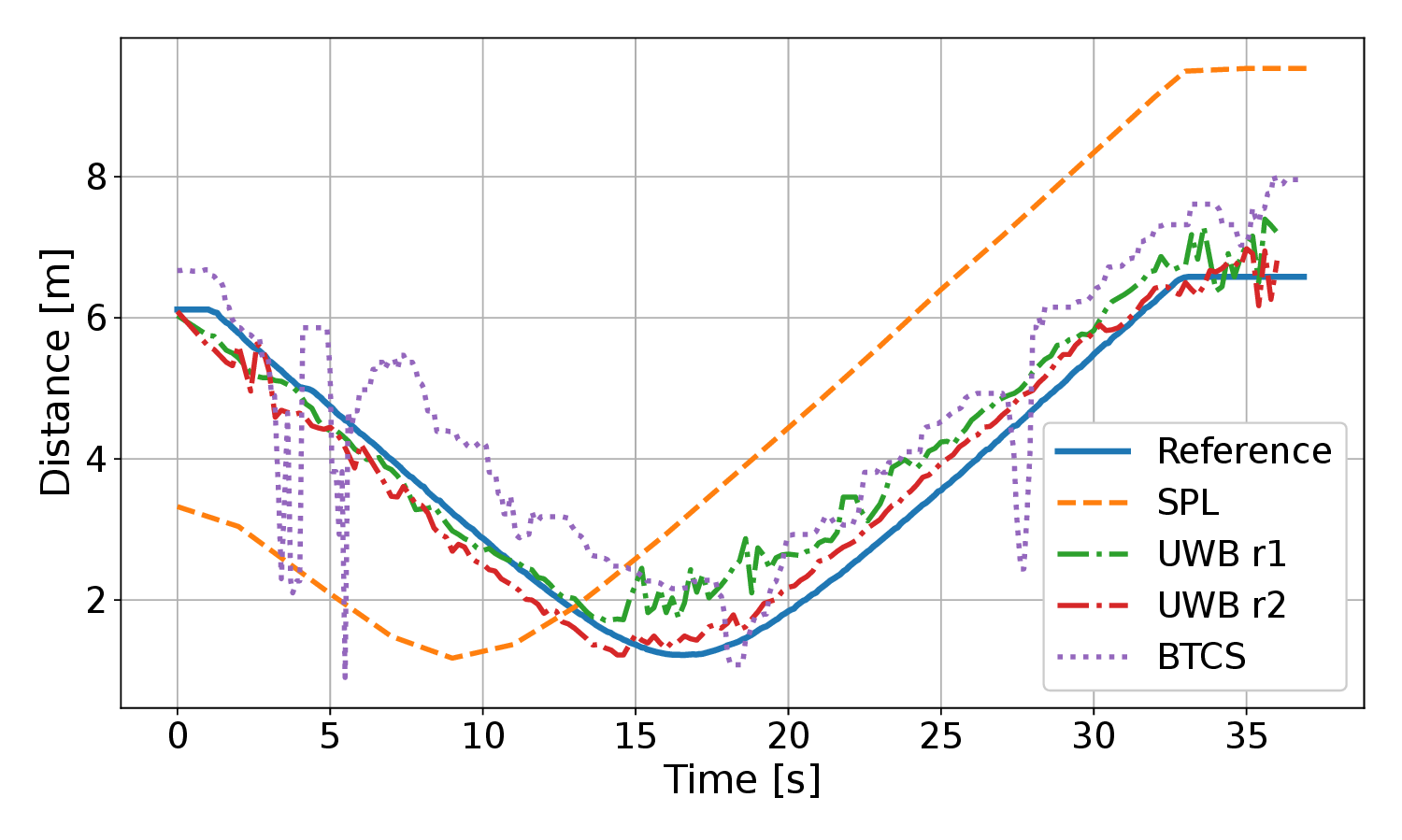}
                \caption{Distance measurements at $l_1 = 1.2m$ with NLOS}
     \end{subfigure} 
\begin{subfigure}[b]{0.44\linewidth}
        \centering
                \includegraphics[width=\textwidth]{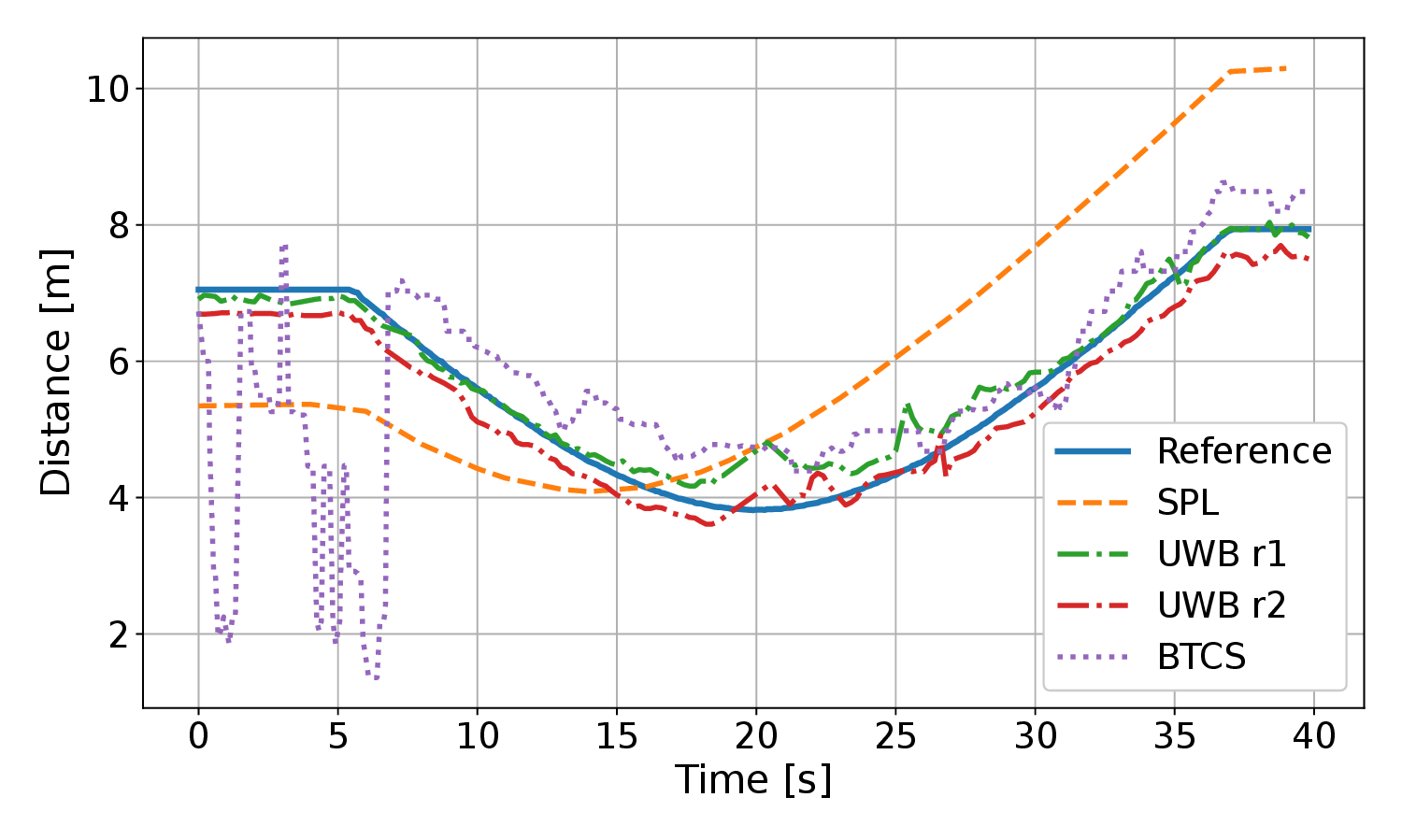}
                \caption{Distance measurements at $l_2 = 3.7m$ with NLOS}
     \end{subfigure} 
    \caption{Figure showing the variation of distance measurements from various localization modalities in an open area.}
    \label{fig:dist_wideopen}
     \end{figure*}

\begin{table}[h]
\caption{Distance measurements in meters in meters from an experiment conducted near a built up area}
\label{tab:builtup}
\resizebox{\linewidth}{!}{
\begin{tabular}{p{\x\linewidth}p{\x\linewidth}p{\x\linewidth}p{\x\linewidth}p{\x\linewidth}p{\x\linewidth}p{\x\linewidth}p{\x\linewidth}p{\x\linewidth}}
 & \multicolumn{2}{c}{$P_{1}$} & \multicolumn{2}{c}{$P_{2}$} & \multicolumn{2}{c}{$P_{1}^{occ}$} & \multicolumn{2}{c}{$P_{2}^{occ}$}\\
 & RMSE  & StD   & RMSE  & StD   & RMSE  & StD   & RMSE  & StD \\ 
\hline
SPL: & 17.307 & 2.939 & 20.986 & 1.492 & 18.271 & 1.437 & 17.68 & 0.976 \\ 
UWB: & 0.507 & 0.458 & 0.422 & 0.386 & 1.958 & 0.695 & 0.532 & 0.52  \\ 
BLE: & 1.39  & 1.076 & 1.906 & 1.397 & 2.75  & 0.959 & 2.388 & 1.836 \\ 
\hline
\end{tabular}
}
\end{table}

\begin{table}[h]
\caption{Distance measurements in meters from an experiment conducted near a built up area under a roof with limited LOS to the GNSS satellites}
\label{tab:roof}
\resizebox{\linewidth}{!}{
\begin{tabular}{p{\y\linewidth}p{\y\linewidth}p{\y\linewidth}p{\y\linewidth}p{\y\linewidth}p{\y\linewidth}p{\y\linewidth}p{\y\linewidth}p{\y\linewidth}}
& \multicolumn{2}{c}{$P_{1}$} & \multicolumn{2}{c}{$P_{2}$} & \multicolumn{2}{c}{$P_{1}^{occ}$} & \multicolumn{2}{c}{$P_{2}^{occ}$} \\
 & RMSE  & StD   & RMSE  & StD   & RMSE  & StD   & RMSE  & StD  \\
\hline
SPL: & 20.762 & 4.111 & 22.084 & 3.94 & 20.05 & 4.436 & 22.202 & 3.365 \\
UWB: & 0.626 & 0.613 & 0.493 & 0.493 & 0.959 & 0.911 & 0.472 & 0.452 \\
BLE: & 2.012 & 1.21  & 2.46  & 1.773 & 2.413 & 1.88  & 6.483 & 4.977 \\
\hline
\end{tabular}
}
\end{table}

 \begin{figure*}[h]
    \centering
    
    \begin{subfigure}[b]{0.44\linewidth}
        \centering
        \includegraphics[width=\textwidth]{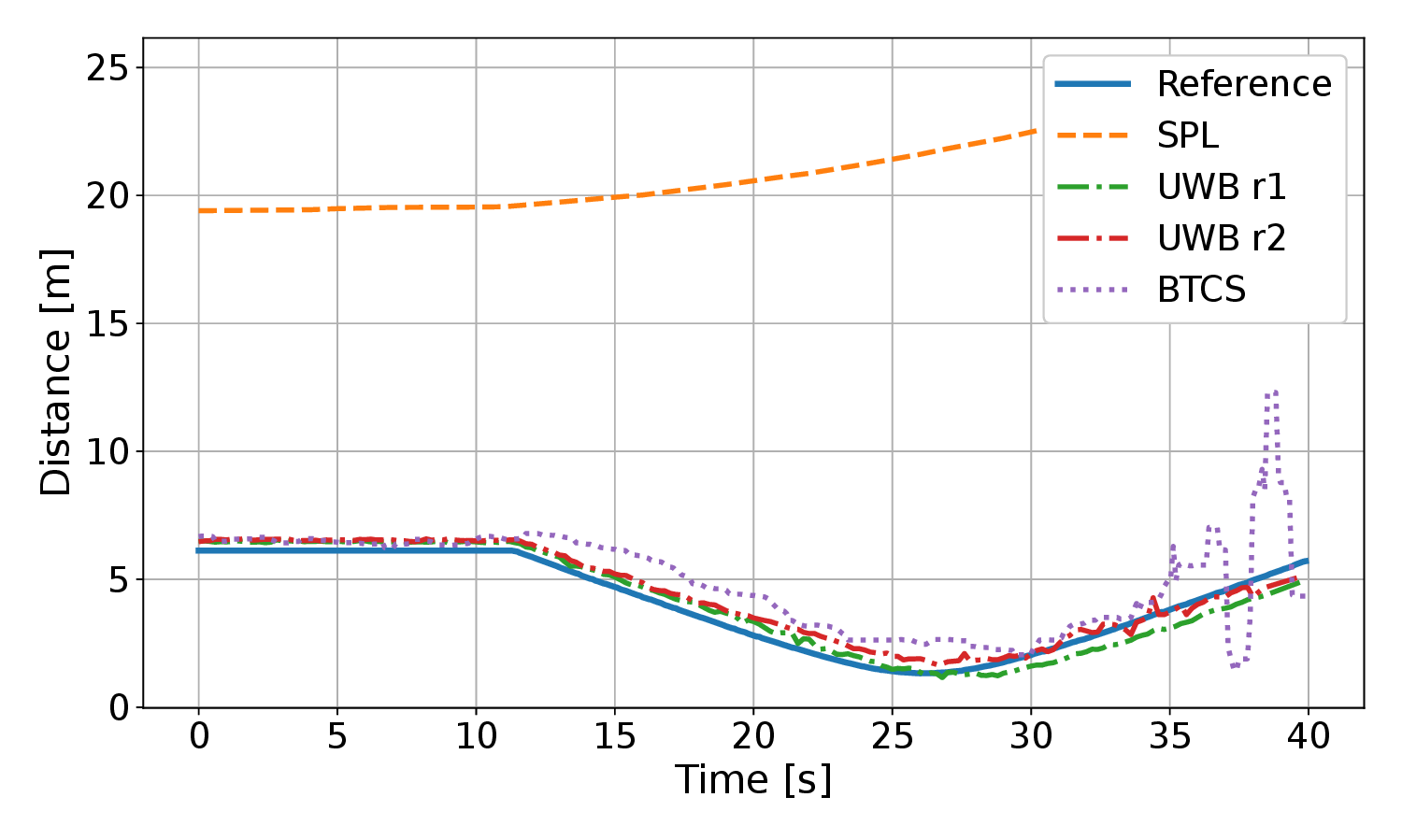}
        \caption{Distance measurements at $l_1 = 1.2m$ with LOS}
     \end{subfigure} 
         \begin{subfigure}[b]{0.44\linewidth}
        \centering
                \includegraphics[width=\textwidth]{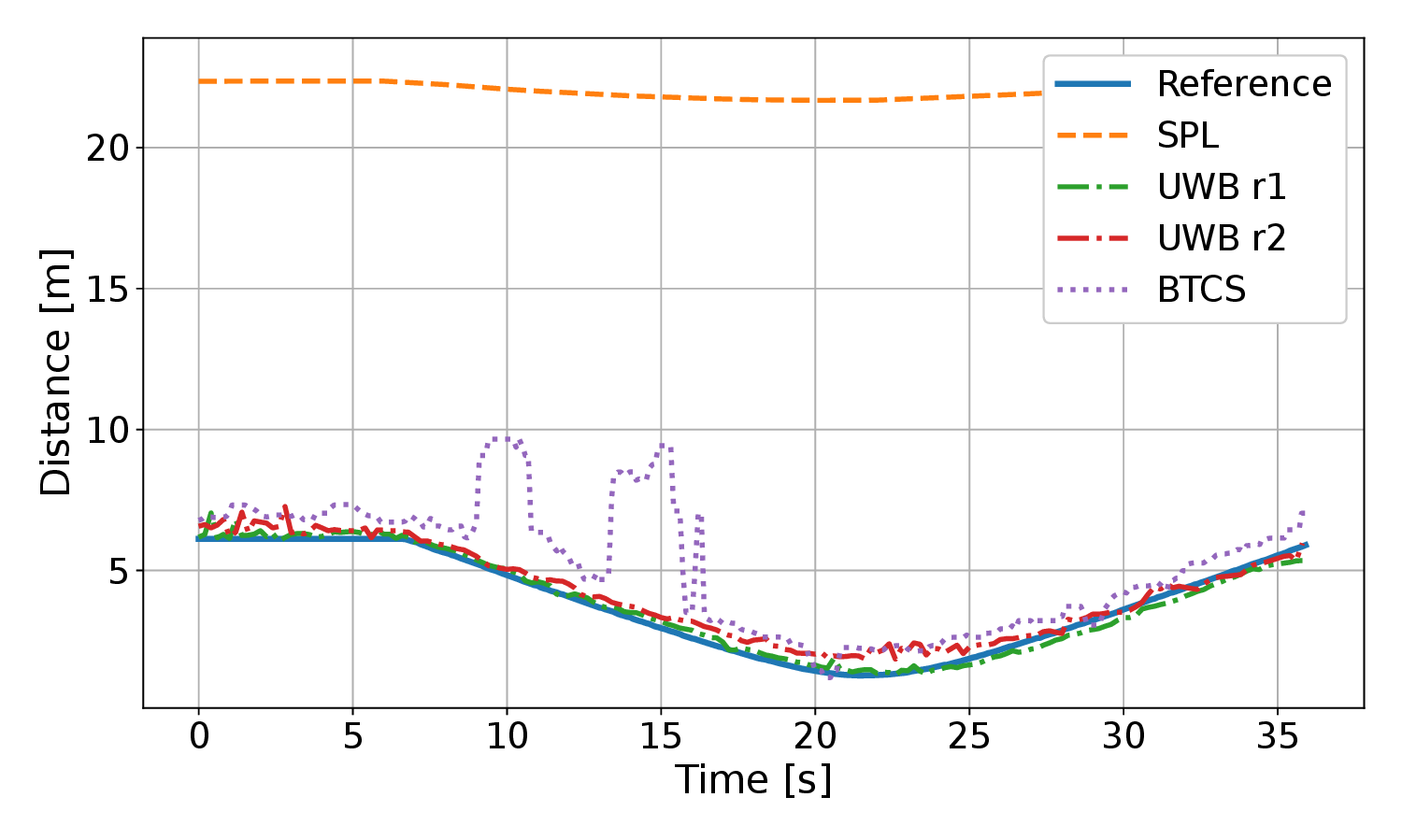}
                \caption{Distance measurements at $l_1 = 1.2m$ with NLOS}
     \end{subfigure} 
     \caption{Figure showing the variation of distance measurements from various localization modalities in a built up area.}
     \label{fig:dist_builtup}
     \end{figure*}
     
     \begin{figure*}[h]
    \centering
    
    \begin{subfigure}[b]{0.44\linewidth}
        \centering
        \includegraphics[width=\textwidth]{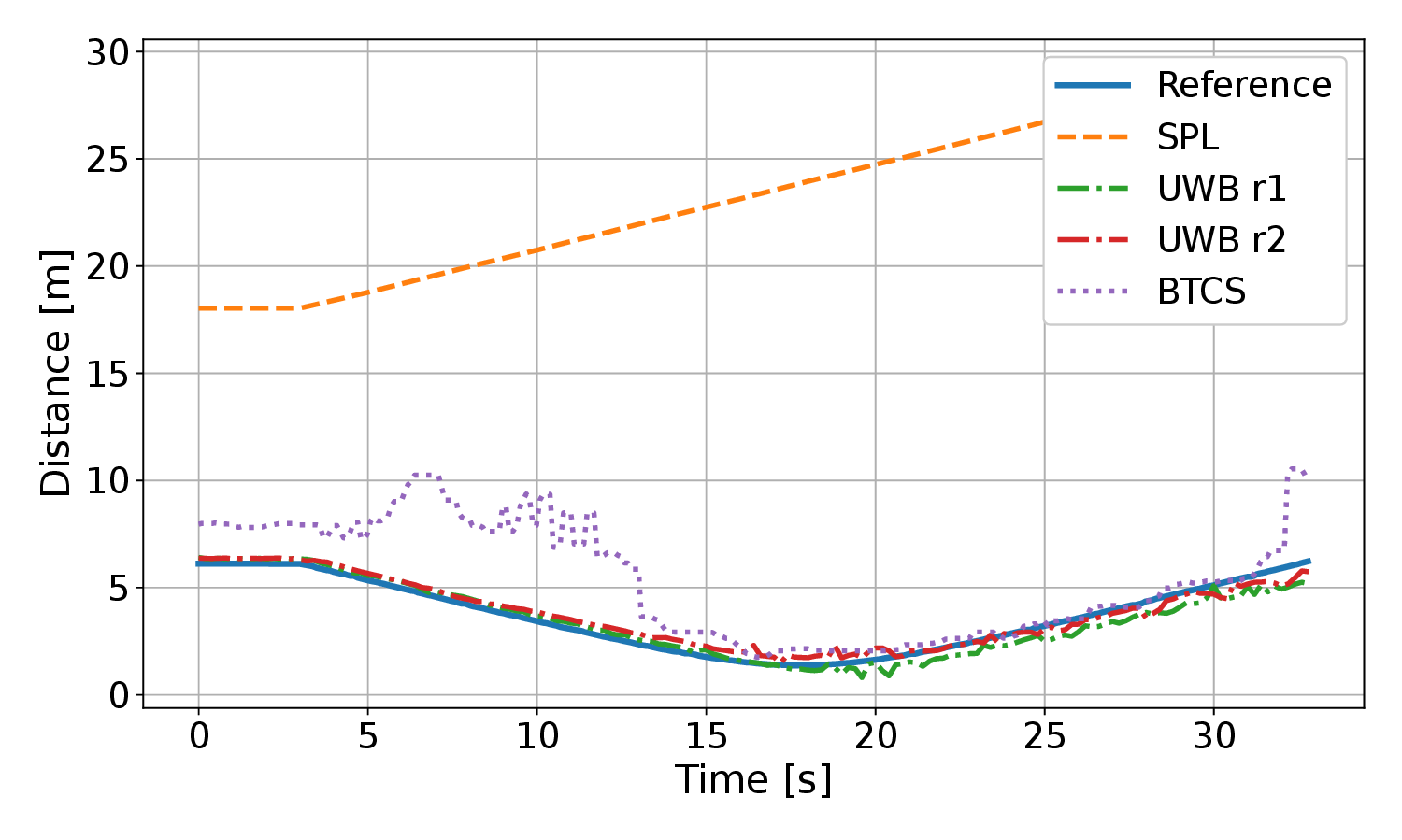}
        \caption{Distance measurements at $l_1= 1.2m$ and NLOS}
     \end{subfigure} 
         \begin{subfigure}[b]{0.44\linewidth}
        \centering
                \includegraphics[width=\textwidth]{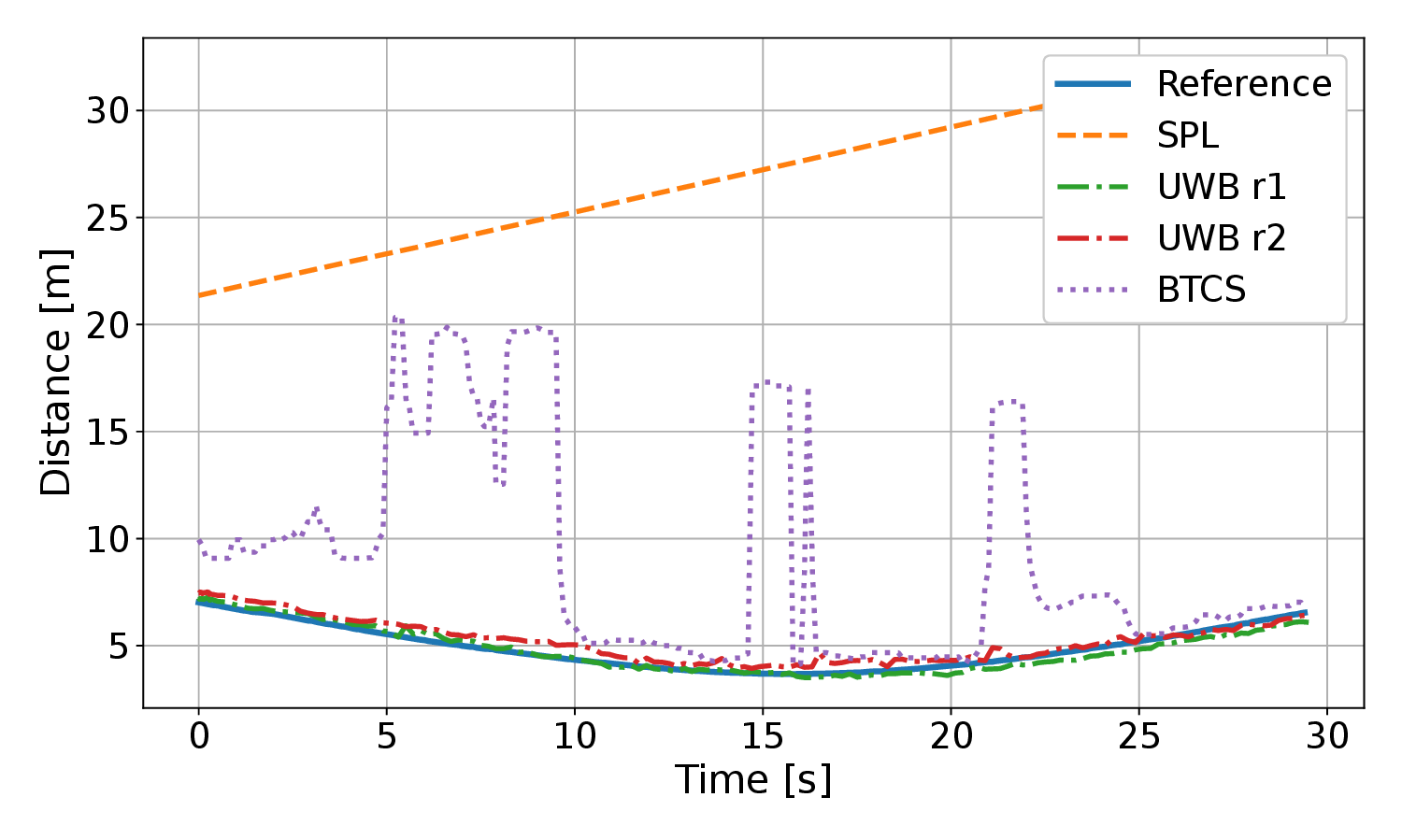}
                \caption{Distance measurements at $l_2 = 3.7m$ and NLOS}
     \end{subfigure} 
          \caption{Figure showing the variation of distance measurements from various localization modalities under a roofed area.}
          \label{fig:dist_roofed}

     \end{figure*}

\begin{figure*}[h]
    \centering
         \begin{subfigure}[b]{0.32\linewidth}
        \centering
                \includegraphics[width=\textwidth]{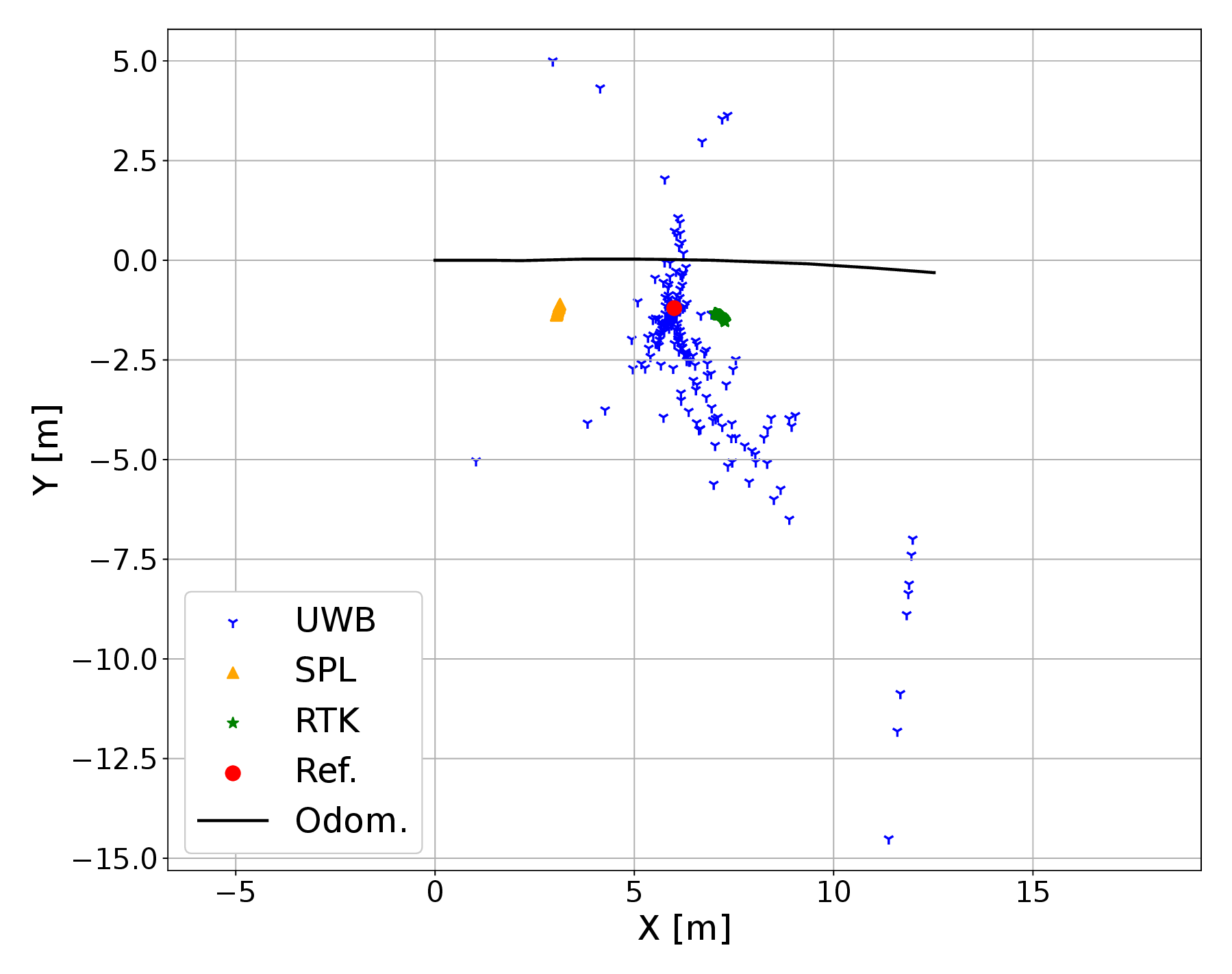}
                \caption{Wide open}
     \end{subfigure} 
    \begin{subfigure}[b]{0.32\linewidth}
        \centering
        \includegraphics[width=\textwidth]{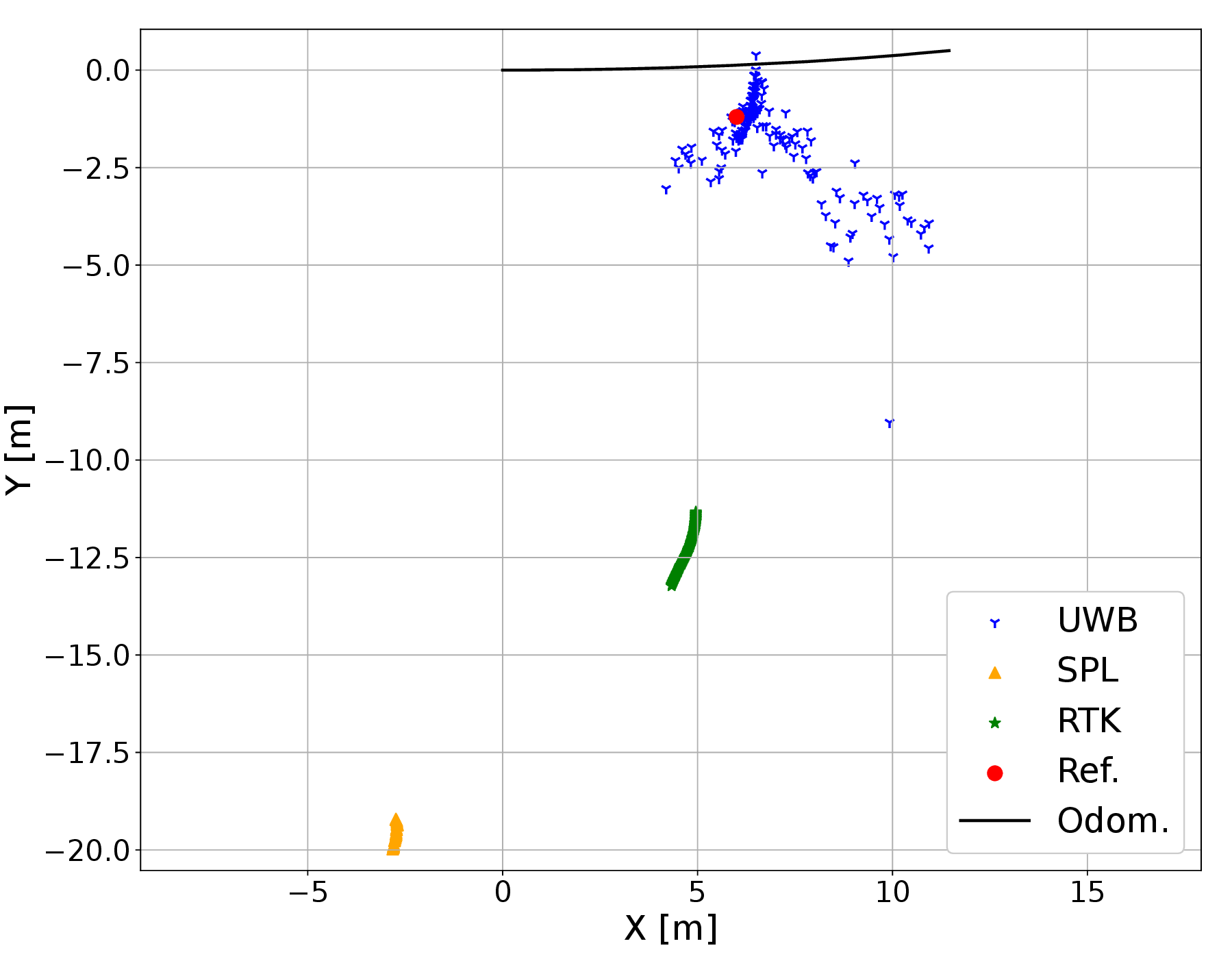}
        \caption{Near a building}
     \end{subfigure} 
\begin{subfigure}[b]{0.32\linewidth}
        \centering
                \includegraphics[width=\textwidth]{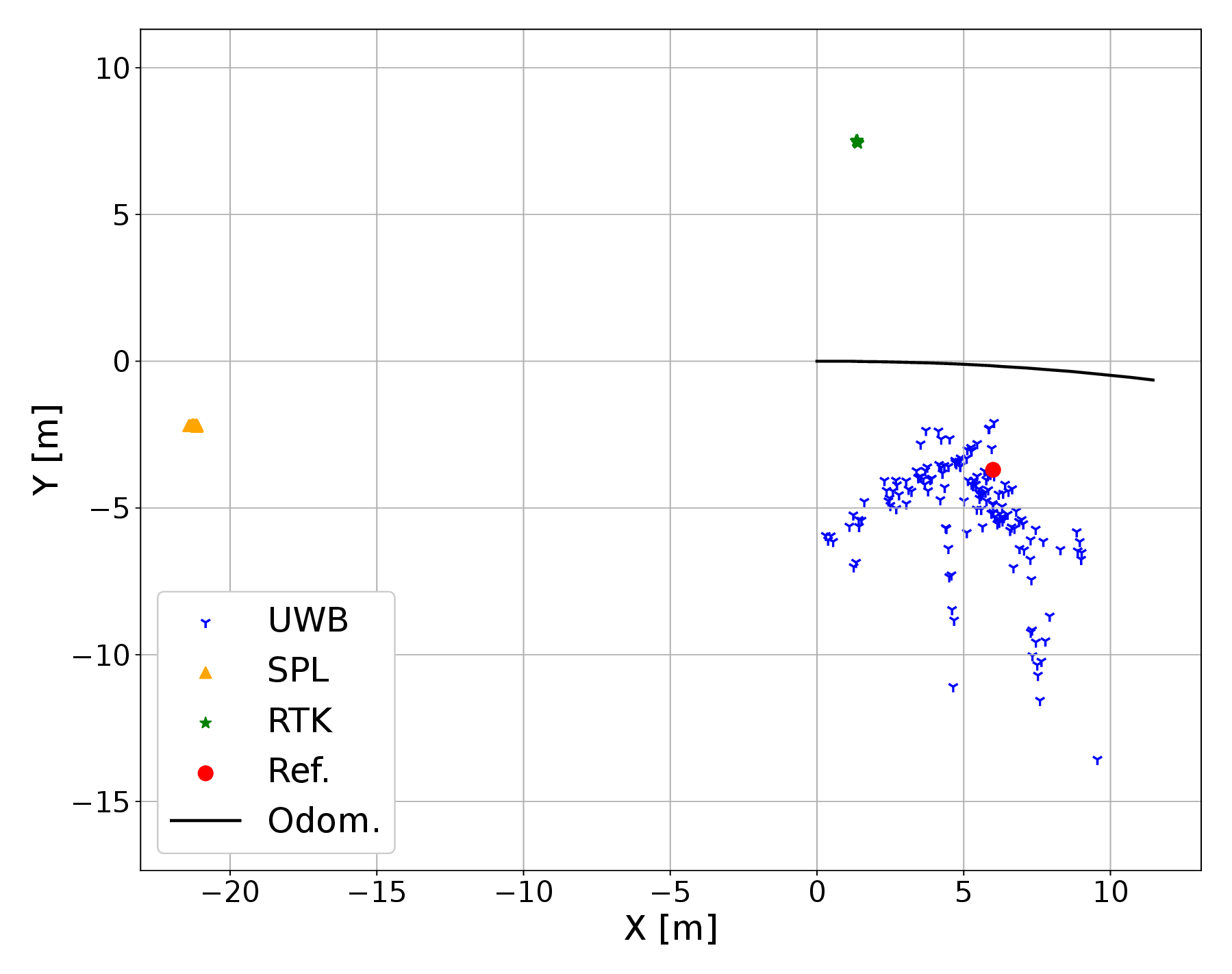}
                \caption{Under a roof}
     \end{subfigure} 
    \caption{Plot showing reference position of the pedestrian in cartesian coordinate frame and as perceived by RTK-corrected GNSS, UWB and SPL over the course of the experiment. The paths of the robot in each experiment are also shown. RTK shows the pedestrian position when measured using an RTK-corrected GNSS module. ``Ref'' refers to the theoretical positions of the pedestrian with respect to the the theoretical ``Start'' coordinate as the origin.}
    \label{fig:xy}
     \end{figure*}

The results in \autoref{tab:open} show that UWB achieves the lowest RMSE and standard deviation in all cases, indicating the most stable performance in open-field conditions. BTCS, while also maintaining low ranging errors, exhibits noticeable degradation when NLOS configurations are introduced. In these cases, reflected and obstructed paths introduce additional propagation delay and multipath effects. However, due to its larger bandwidth and fine temporal resolution, UWB is generally more resilient to these effects than BTCS. GNSS shows larger distance errors, especially in the short distance and LOS scenario. However, for GNSS, this pedestrian–robot distance or LOS/NLOS conditions do not directly influence the behavior. Instead, occasional coverage degradations lead to significant localization errors.

In the condition in which the experiment was carried out near buildings (see \autoref{tab:builtup}) or was covered by a roof structure to partially attenuate satellite and radio signal propagation (see \autoref{tab:roof}), the performance of UWB and BTCS maintain a similar behavior, although a slight increase in RMSE and standard deviation, particularly for BTCS, can be observed. This degradation can be attributed to multipath effects generated by nearby buildings, or the roof itself. However, GNSS performance remains consistently inaccurate in these scenarios, where coverage is limited, even though it is still operating outdoors.


Distance measurements from different localization sensors for all experiments were analyzed. UWB performs consistently better than all the other modules in all three scenarios, even under NLOS conditions, as shown in \autoref{fig:dist_wideopen}, \autoref{fig:dist_builtup} and \autoref{fig:dist_roofed}. UWB distance measurements from two anchors are shown as ``UWB r1'' and ``UWB r2'' in the figures mentioned above. BTCS also has an accurate ranging performance, but it shows an increase in jitter under NLOS conditions, as seen in \autoref{fig:dist_builtup}. This effect is amplified under the roof, as observed in \autoref{fig:dist_roofed}, where signal reflections from the roof structure increase multipath effects. These figures further highlight the limitations of GNSS based SPL, which is unable to provide meter-level ranging accuracy, even in open-field conditions. Please note that the ``Reference'' mentioned in the graphs is calculated by using odometry provided by a Kalman filter onboard the robot.

Changes in the pedestrian–robot distance do not significantly affect the ranging error of any technology, since all measurements were performed within the transmission range of the devices. However, as shown in \autoref{fig:dist_roofed}, in NLOS conditions, BTCS exhibited more frequent and higher deviations at larger distances. This suggests that while the communication link was maintained, the ranging stability of BTCS can still be degraded when the distance is increased, especially in NLOS scenarios.

While UWB demonstrates the highest ranging precision among the evaluated sensors, converting multiple anchor-based distance measurements into a position estimate introduces additional errors. In multilateration, these individual ranging errors accumulate and manifest as a spatial dispersion of the pedestrian’s estimated location. This effect is illustrated in \autoref{fig:xy}, where the estimated position of the pedestrian based on UWB is distributed around the true location. GNSS, in contrast, provides comparatively stable positions, but suffers from a substantial translational offset from the true position, making it unsuitable for safety-critical applications. The limitations of GNSS become even more pronounced near buildings or under roof coverage, where satellite signals are affected by NLOS, and this obstruction further degrades the performance. Even with RTK corrections, reliable high-precision positioning cannot be consistently achieved in such environments.

\section{Discussion}\label{sec:discussion}

From \autoref{tab:open}, \autoref{tab:builtup} and \autoref{tab:roof}, it can be observed that all communication modalities perform better in the wide open area. The degradation in performance is especially noticeable when positioning relies on the smartphone’s built-in GNSS. Even with RTK corrections, the GNSS positioning near buildings was inaccurate, leading to position and orientation errors. The degradation in performance of UWB and BTCS was within reasonable limits, leading them to be strong candidates for warning VRUs of dangerous situations. Since these risky situations are more likely to occur in cities where there are a higher density of buildings, the results favor UWB and BTCS. Among these two technologies, UWB performs better than BTCS for localization. 

It should also be noted that the experiments were conducted using development kits for UWB and BTCS, which may not directly reflect large-scale deployment conditions. While GNSS is almost universally available in commercial devices, UWB and BTCS depend on hardware support in newer smartphone generations. UWB is increasingly standardized through the FiRa Consortium, but it typically relies on another communication technology, such as Bluetooth, for device discovery and pairing. This dependency limits scalability of UWB and restricts native communication mainly to ranging data. Therefore, additional channels (e.g., BLE) would be required to transmit warning messages in practical safety applications. 

The ubiquity of UWB and BTCS technology in modern smartphones presents a further challenge. UWB adoption is currently limited in smartphones. This could become a problem so difficult that it could  prevent the use of UWB as a communication technology for VRU warning systems. In contrast, Bluetooth 6.0, which includes BTCS, is on track to be widely adopted by smartphone manufacturers which includes the BTCS feature. Therefore BTCS might gain relevance in the following years as a technology that will be used for pedestrian localization and communication. 

\section{Conclusion}\label{sec:conclusion}

This study evaluated the ranging and positioning performance of UWB, BTCS, and GNSS in open, built-up, and roofed environments. The results show that UWB achieves the highest ranging precision and stability, maintaining low RMSE and standard deviation across all scenarios, even under NLOS conditions. BTCS generally provides accurate ranging, but performance degrades under NLOS and at larger pedestrian–robot distances, particularly in multipath-prone environments such as near buildings or under roof coverage. GNSS, while widely available in commercial devices, demonstrates poor positioning accuracy in all tested environments, with substantial errors even in open areas and severe degradation in built-up or roofed scenarios.

While UWB offers high accuracy, its integration into commercial devices is limited due to requirements for an additional communication technology for device discovery and message exchange. BTCS, in contrast, is expected to be widely supported in upcoming smartphones, making it a practical alternative or complement to UWB. Therefore, a hybrid approach combining UWB and BTCS could take advantage of the accuracy of UWB and the broader device compatibility of BTCS, while GNSS alone is insufficient for meter-level local positioning. Future work should investigate sensor fusion approaches that combine UWB and BTCS with complementary modalities, such as camera-based systems or GNSS, to improve robustness in realistic and cluttered environments. Targeted algorithms for UWB and BTCS to improve localization accuracy can also improve positioning accuracy when using multiple anchors as mentioned in \cite{kim2025deep} and \cite{park2024pedloc}.

\section*{ACKNOWLEDGMENTS}
This work has been conducted as a part of OptiPEx project (No.101146513) funded by the European Union. Views and opinions expressed are however those of the author(s) only and do not necessarily reflect those of the European Union or CINEA. Neither the European Union nor the granting authority can be held responsible for them.


%
\bibliographystyle{IEEEtran}
\bibliography{ref}

\end{document}